\documentclass[letterpaper, 10 pt, conference]{ieeeconf} 
\IEEEoverridecommandlockouts                            
\usepackage{cite}
\usepackage{amsmath,amssymb,amsfonts}
\usepackage{algorithm}
\usepackage{graphicx,import}
\usepackage{textcomp}
\usepackage{xcolor}
\usepackage{bm,nicefrac}
\usepackage{mathrsfs}
\usepackage{algpseudocode}
\usepackage{multirow}
\usepackage{url}
\def\BibTeX{{\rm B\kern-.05em{\sc i\kern-.025em b}\kern-.08em
    T\kern-.1667em\lower.7ex\hbox{E}\kern-.125emX}}
\usepackage{array}
\newcolumntype{L}[1]{>{\raggedright\let\newline\\\arraybackslash\hspace{0pt}}m{#1}}
\newcolumntype{C}[1]{>{\centering\let\newline\\\arraybackslash\hspace{0pt}}m{#1}}
\newcolumntype{F}[1]{>{\let\newline\\\arraybackslash\hspace{0pt}}m{#1}}
\definecolor{tudelft-fuchsia}{cmyk}{0.19,1,0,0.19}
\definecolor{tudelft-cyan}{cmyk}{1,0,0,0}

\newtheorem{assumption}{Assumption}
\newtheorem{rem}{Remark}

\title{\LARGE \bf  Fault-tolerant Control of Robot Manipulators with Sensory Faults using Unbiased Active Inference}

\author{Mohamed Baioumy$^{1*}$,  Corrado Pezzato$^{2*}$, Riccardo Ferrari$^{2}$, Carlos Hern\'andez Corbato$^{2}$  and Nick Hawes$^{1}$
\thanks{{*These authors contributed equally to this work.}}
\thanks{$^{1}$ Authors are with the Oxford Robotics Institute, University of Oxford. For correspondence \{mohamed, nickh\}@robots.ox.ac.uk}%
\thanks{$^{2}$ Authors are with Delft University of Technology. For correspondace \{c.pezzato, c.h.corbato, r.ferrari\}@tudelft.nl}%
\thanks{This work was supported by Ahold Delhaize}%
}

\begin{document}

\maketitle
\thispagestyle{empty}
\pagestyle{empty}

\begin{abstract}

This work presents a novel fault-tolerant control scheme based on active inference. Specifically, a new formulation of active inference which, unlike previous solutions, provides unbiased state estimation and simplifies the definition of probabilistically robust thresholds for fault-tolerant control of robotic systems using the free-energy. The proposed solution makes use of the sensory prediction errors in the free-energy for the generation of residuals and thresholds for fault detection and isolation of sensory faults, and it does not require additional controllers for fault recovery. Results validating the benefits in a simulated 2-DOF manipulator are presented, and future directions to improve the current fault recovery approach are discussed.


\end{abstract}

\section{Introduction}
Fault tolerant (FT) control is of vital importance for many applications such robots working in production lines \cite{kruger}. In order to guarantee high standards of reliability and security, the area of FT control of robot manipulators increasingly gathered importance in recent years \cite{surveyFT}. Many approaches have been developed for faults in actuators and sensors \cite{actFault1, actFault2, sensorFault1, sensorFault2,  sensorFault3}. Model based fault tolerant techniques are amongst the most advanced approaches to tackle the problem of fault detection and isolation (FDI) for dynamical systems \cite{ChengBook}. These methods rely on monitoring system outputs using mathematical models to generate residual signals which are compared to a threshold. Faults are detected if the threshold is exceeded, while the actions for fault recovery are usually performed through controller re-design or by switching to a different controller \cite{Narendra}. 


Regarding FT control for robot manipulators, \cite{sensorFault2} performs fault diagnosis in image-based visual servoing. The residual signal is defined as the root mean squared estimation error (RMSE) of a Kalman filter, which predicts the values of a number of features in the camera image. The residual is compared against a user-defined static threshold and fault isolation is achieved via a decision table. The controller reconfiguration is not designed in the paper, and it would require an additional element on top of the FDI scheme. Visual information from a camera can also be used to compensate for lack of observability in case of sensory faults. In \cite{sensorFault3}, three different second-order sliding mode observers are used for generating residuals using independent measurements from camera images. The detection thresholds were selected to minimize the probability of false or missed alarms on the basis of simulation and experimental data. However, the visual subsystem is assumed faultless. 

FDI through static thresholds can be effective, but stochastic decision theory has been shown to outperform these methods. The actual distributions of residuals can be estimated such that a user can define thresholds based on their acceptable probability of false positives \cite{fang2015mooring,ferrari2017message,Rostampour2017,rostampour2020privatized}. The work in \cite{blas2011stereo} showed fault-tolerant steering control of an agricultural vehicle for bailing, detecting sensory faults and artefacts in images through statistical learning. The sensors configuration for control was then decided at run-time.

Generally speaking, model-based fault tolerant control schemes rely on additional supervision blocks that increase system complexity. FDI and fault recovery present two main challenges: 
a) The definition of robust yet sensitive pairs of residuals and thresholds;
b) The design of specific additional supervisory systems to accommodate large faults.

The work presented in this paper extends the active inference controller (AIC) recently developed by the authors in \cite{Pezzato2020, baioumy2020active}. Active inference relies on approximate Bayesian inference \cite{friston2, buckley} for prediction errors minimization, and it allows state estimation and control using a single cost function. This scheme showed remarkable capabilities while dealing with missing sensory information or large unmodeled dynamics, see \cite{lanillos1, Pezzato2020} respectively. The authors in \cite{Corrado2} initially presented a fault tolerant control scheme with an AIC which showed great promise. However, this presented a few fundamental limitations which will be thoroughly discussed in Sec.~\ref{sec:AICFDI}. Most importantly, FDI in \cite{Corrado2} can be affected by false-positives when target state changes, and it relies on a static threshold which might be over-conservative. In this work, we: improve the state estimation of a standard AIC; develop an unbiased AIC (u-AIC); reduce the probability of false-positives and allow to easily define a probabilistically robust threshold for fault detection.

The paper is organised as follow: Sec. \ref{sec:preliminaries} presents the background on active inference for fault-tolerant control highlighting the limitations of past work. Sec. \ref{sec:decoupled} presents a new formulation of the AIC with unbiased state estimation, while Sec. \ref{sec:FDI} explains how fault tolerant control is achieved with this new formulation. Results for a simulated 2-DOF robot arm are presented in Sec. \ref{sec:results} while Sec. \ref{sec:conclusions} reports a summary and future challenges.

\section{Preliminaries}
\label{sec:preliminaries}
This section presents the background knowledge on active inference for robot control \cite{Pezzato2020, Corrado2} required to understand and justify the need for an unbiased AIC.

\subsection{Active inference controller (standard AIC)}
\label{sec:problemStatement}
\begin{figure}[!htb]
    \centering
    \includegraphics[width=0.45\textwidth]{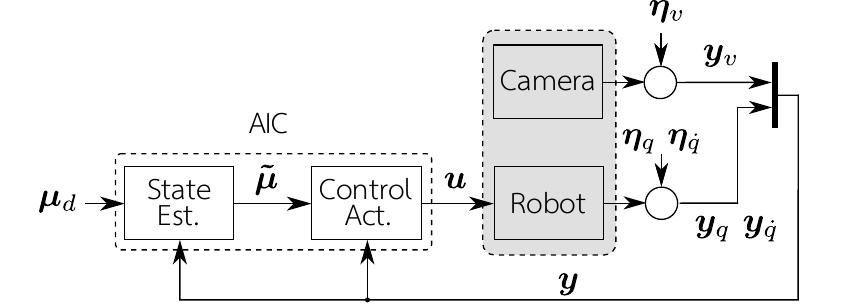}
    \caption{General AIC control scheme for joint space torque control.}
    \label{fig:generalscheme}
\end{figure}

The AIC in \cite{Pezzato2020, Corrado2} is defined for torque control in joint space of a generic robot manipulator (Fig.~\ref{fig:generalscheme}). The robot arm is equipped with encoders and velocity sensors with relative sensory readings $\bm y_q,\ \bm y_{\dot{q}}$. In addition, the end-effector Cartesian position $\bm y_v$ is made available through a camera. The system's output is represented by $\bm y = [\bm y_q,\ \bm y_{\dot{q}},\ \bm y_v]\in\mathbb{R}^d $. The proprioceptive sensors and the camera are affected by zero mean Gaussian noise $\bm \eta=[\bm \eta_q$, $\bm \eta_{\dot{q}}$, $\bm \eta_v]$. The camera is affected by barrel distortion. The AIC can be used to control the robot arm to a desired configuration $\bm \mu_d$, providing the control input $\bm u\in\mathbb{R}^m$ as torques to the single joints. To this end, the control input is computed on the basis of the so-called \emph{free-energy} and the \emph{generalised motions} $\tilde{\bm \mu}$ \cite{Corrado2}.

The free-energy principle relies on Bayesian inference \cite{Bayes} for state estimation. 
The goal is to find the posterior over states given observations, $p(\bm{x}| \bm y_v, \bm y_q, \bm y_{\dot{q}})$. This computation is in general intractable using Bayes' rule and in practice the true posterior distribution is approximated with a variational distribution $Q(\bm{x})$ \cite{fox2012tutorial}. The most probable state is computed by minimizing the Kullback-Leibler divergence ($D_{KL}$) between $p(\bm{x}|\bm{y})$ and $Q(\bm x)$ \cite{buckley}:
\begin{align}
\nonumber 
    D_{KL}(Q(\bm x)||p(\bm{x}|\bm{y})) = & \int  Q(\bm x) \ln\frac{Q(\bm x)}{p(\bm{x}|\bm{y})}d\bm x \\
    = & F+\ln p(\bm y) 
\end{align}
where $F$ represents the \emph{free-energy}. By minimizing $F$, $Q(\bm{x})$ approaches the true posterior $p(\bm{x}|\bm{y})$. The probabilistic model $p(\bm{x}, \bm y_v, \bm y_q, \bm y_{\dot{q}})$ is factorized as: 
\begin{equation}
  p(\bm{x}, \bm y_v, \bm y_q, \bm y_{\dot{q}}) =  \underbrace{p(\bm y_v|\bm{x}) p(\bm y_q|\bm{x}) p(\bm y_{\dot{q}}|\bm{x})}_{observation\hspace{1mm} model} \underbrace{p(\bm{x})}_{prior}
  \label{eq: factorization_of_coupled_AIC}
\end{equation}

If $Q(\bm{x})$ is a Gaussian with mean $\bm{\mu}$ and the Laplace approximation is utilized \cite{variational}, then $F$ reduces to 
\begin{equation}
\label{eq:F_prob}
    F = -\ln p(\bm{\mu}, \bm y_v, \bm y_q, \bm y_{\dot{q}}) + C \, ,
\end{equation}

\noindent where $C$ is a constant. Higher order derivatives of the state $\bm{x}\in\mathbb{R}^n$ are combined in the term $\bm{\tilde{{\mu}}}$  (i.e. $\bm{\tilde{\mu}} = [\bm{\mu}, \bm{\mu'}, \bm{\mu''}]$). 
This is referred to as \emph{generalized motions} \cite{buckley, friston1}. The number of generalised motions considered will affect the update laws for state estimation and control. 

In order to characterize the free-energy and compute it, one has to define two generative models, one for the state dynamics, and one for the observations. The latter is modeled according to \cite{buckley} as $\bm y = \bm g(\bm \mu) +\bm z$, where $\bm g(\bm \mu) = [\bm g_{q},\ \bm g_{{\dot{q}}},\ \bm g_{v}] :\mathbb{R}^d \mapsto \mathbb{R}^d$ represents the non-linear mapping between sensory data and states of the environment, and $\bm z$ is Gaussian noise $\bm z \sim (\bm 0,\Sigma_y)$. The generative model of the state dynamics is defined as\cite{buckley}:
\begin{equation}
	\frac{d\bm{\mu}}{dt}=\bm \mu' = \bm{f}(\bm{\mu})+\bm{w}
\end{equation} 
where $\bm f(\bm \mu):\mathbb{R}^n \mapsto \mathbb{R}^n$ is a generative function dependant on the belief about the states $\bm{\mu}$ and $\bm w$ is Gaussian noise $\bm w \sim (\bm 0,\Sigma_\mu)$. As in previous work \cite{baioumy2020active}, we define $\bm f(\cdot)$ such that the robot will be steered to a desired joint configuration $\bm \mu_d$ following the dynamics of a first order linear system with time constant $\tau$. 
\begin{equation}
    \label{eq:genModmu}
	\bm{f}(\bm{\mu}) = (\bm \mu_d -\bm \mu)\tau^{-1}
\end{equation}

From equation \eqref{eq:genModmu} we can see that the desired state is encoded in the prior. The time constant $\tau$ influences the generative model of the state dynamics $\bm f(\cdot)$. As explained in \cite{baioumy2020active}, the AIC has two extremes depending on the value of $\tau^{-1}$. If $\tau^{-1} \rightarrow  0$, the AIC only performs filtering and no control. On the other hand, if $\tau^{-1} \rightarrow  \infty$ the AIC is equivalent to a PID controller \cite{baioumy2020active,baltieri2018probabilistic}. For any value in between, there is a compromise between estimation and control. The estimation and control are thus `coupled' and  state-estimation with the standard AIC results in a bias towards the desired target, see \eqref{eq:F_minimize_mu}. If all distributions in equation \eqref{eq: factorization_of_coupled_AIC} are assumed Gaussian, the expression for free-energy simplifies to (see \cite{Pezzato2020} for complete derivations):
\begin{align}
        \label{eq:free-energy}
    	F = \frac{1}{2}\sum_i\begin{pmatrix} \bm\varepsilon_i^\top P_{i} \bm\varepsilon_i - \ln{|P_{i}|}\end{pmatrix} + C,
\end{align}
where $i\in~\{y_q,\ y_{\dot{q}},\ y_v,\ \mu,\ \mu'\}$, and $P_{i}$ defines a precision (or inverse covariance) matrix. Note that we set $\tau = 1$ as in \cite{oliver, Pezzato2020}. The terms $\bm\varepsilon_i$ with $i\in \{y_q,\ y_{\dot{q}},\ y_v\}$ are the \emph{Sensory Prediction Errors} (SPE), representing the difference between observed sensory input and expected one. \textcolor{black}{In general, the SPE for a sensor $s$ are defined as $(\bm y_s - \bm g_s(\bm\mu))$.} The model prediction errors are instead defined considering the desired state dynamics as $\bm\varepsilon_{\mu} = (\bm \mu' - \bm f(\bm\mu))$ and $\bm\varepsilon_{\mu'} = (\bm \mu'' - \nicefrac{\partial \bm f(\bm\mu)}{\partial \bm \mu} \mu')$. In particular, for the robot arm used in this paper, $\bm\varepsilon_{y_q} = (\bm{y}_{q}-\bm{\mu})$, $\bm\varepsilon_{y_{\dot{q}}} = (\bm y_{\dot{q}}-\bm{\mu}')$, $\bm\varepsilon_v = (\bm{y}_v-\bm{g}_{v}(\bm{\mu}))$, and $\bm \varepsilon_\mu= (\bm{\mu}'+\bm{\mu}-\bm{\mu}_d)$, $\bm \varepsilon_{\mu'}=(\bm{\mu}''+\bm{\mu}')$.  For more details on the derivation of equation \eqref{eq:free-energy}, refer to \cite{Pezzato2020,baioumy2020active,buckley}. 

To achieve state-estimation a gradient descent on the free-energy is used. The variable $\bm{\tilde \mu}$ is updated as:
\begin{equation}
    \label{eq:F_minimize_mu}
    \bm{\dot{\tilde{\mu}}} = D\bm{{\tilde{\mu}}} - \kappa_{\mu}\frac{\partial F}{\partial {\bm{\tilde{\mu}}}},
\end{equation}
\noindent where $\kappa_{\mu}$ is the gradient descent step size, and $D$ is a shifting operator with ones on the upper diagonal. This form of gradient descent is needed due to using generalized motions as mentioned earlier \cite{friston2}. 
The control actions, are also computed through gradient descent; however, the expression for $F$ does not include any actions explicitly. The chain rule is then used, assuming an implicit dependency between actions and sensory input.
\begin{equation}
    \label{eq:F_minimize_a}
    \bm{\dot{u}} = - \kappa_{u}\frac{\partial F}{\partial \bm{u}}
    = - \kappa_{u} \frac{\partial F}{\partial {\bm{y}}}  \frac{\partial {\bm{y}}}{\partial \bm{u}},
\end{equation}

\noindent where $\kappa_{u}$ is the gradient descent's step size. The term $\frac{\partial {\bm{y}}}{\partial \bm{u}}$ is often approximated as linear, similarly to~\cite{Pezzato2020, oliver}.

\subsection{Limitations of fault-tolerant control with standard AIC}
\label{sec:AICFDI}
The AIC was used in \cite{Corrado2} for fault tolerant control. In this subsection we aim at highlighting the main limitations of the AIC, which are addressed in Sec.~\ref{sec:decoupled} through the proposed unbiased active inference scheme. The main idea in \cite{Corrado2} was that the SPE in the free-energy could be used as residuals for fault detection, removing the need for more advanced residual generators. Besides, the precision matrices in the free-energy could be used for fault accommodation, since they represent the trust associated to sensory readings. For a faulty sensor, the relative precision was decreased to zero to perform fault recovery. Specifically, the residuals in \cite{Corrado2} are generated by considering the quadratic form of the SPE $\bm \varepsilon_{s}^\top P_{y_s} \bm \varepsilon_{s}$, where $\bm \varepsilon_s = \bm y_s- \bm g_s(\bm \mu)$ with $s \in \{q,\ \dot{q},\ v \}$. A static fault detection threshold $\psi_s$ was chosen as an upper bound on the quadratic terms. 

\begin{figure}[!htb]
    \centering
    \includegraphics[width=0.8\linewidth]{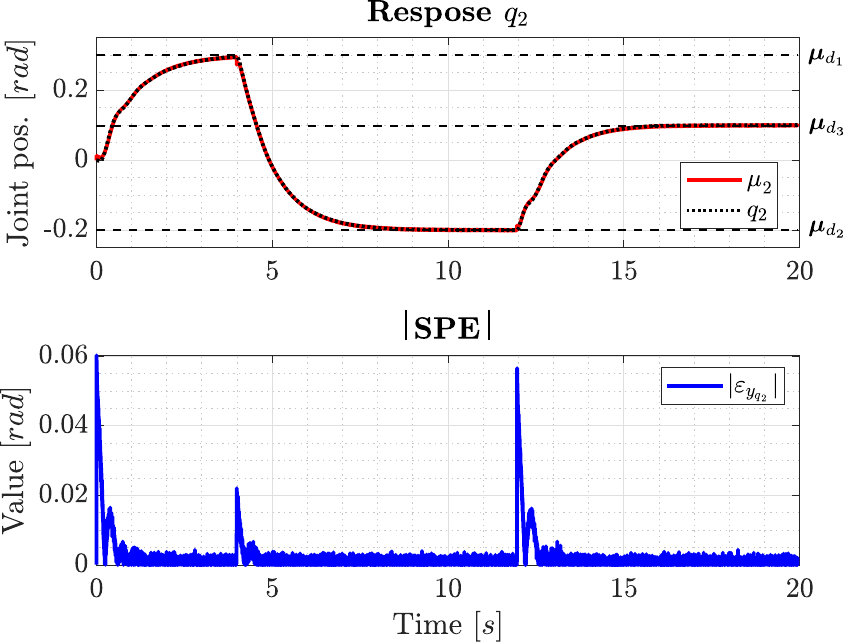}
    \caption{Effect of abrupt change in goal on the SPE in standard AIC. The SPE increases sharply without the presence of any sensory faults.}
    \label{fig:falsePositives}
\end{figure}

The SPE depends on the belief $\bm \mu$, which is biased towards a given goal. This means that the SPE in standard AIC can increase for causes that are not necessarily related to a fault, i.e. the robot is stuck due to a collision or there is an abrupt change in the goal $\bm \mu_d$, for example in a pick-and-place application. The residuals are thus dependent on the goal, which is problematic. This was not an issue in \cite{Corrado2} because of the over-conservative threshold used for fault detection, but it becomes apparent when the SPE are closely analysed. Fig.~\ref{fig:falsePositives} reports the absolute value of the SPE for the joint of a robot arm controlled using an AIC, during a point to point motion with three via-points $\bm \mu_{d_1},\ \bm \mu_{d_2},\ \bm \mu_{d_3}$. 

Using a tighter threshold to detect faults of small entity by using the approach proposed in \cite{Corrado2} would lead to many false positives. In this work we tackle the problem of biased sensory prediction errors and we define an \textit{unbiased AIC} where the SPE are independent from the input, providing more accurate state estimation. With the u-AIC is also possible to use the sensory prediction error directly for fault detection using a probabilistically robust threshold instead of a static one as in \cite{Corrado2}.


\section{Unbiased active inference}
\label{sec:decoupled}
To tackle the problems discussed in the previous section, the unbiased AIC is developed (u-AIC). 
\subsection{Derivation of the u-AIC}
Consider a probabilistic model with explicit actions $p(\bm{x}, \bm{u}, \bm y_v, \bm y_q, \bm y_{\dot{q}})$, where unlike the AIC, $\bm{x} = [\bm q \,\, \dot{\bm q}]^\top$. The distribution factorizes as: 
\begin{equation}
    p(\bm{x}, \bm{u}, \bm y_v, \bm y_q, \bm y_{\dot{q}}) = \underbrace{p(\bm{u}|\bm{x})}_{control} \underbrace{p(\bm y_v|\bm{x}) p(\bm y_q|\bm{x}) p(\bm y_{\dot{q}}|\bm{x})}_{observation \hspace{1 mm}model} \underbrace{p(\bm{x})}_{prior} 
    \label{eq: factorization_d-aic_joint_probabilistic_distribution_ with_explicit_actions}
\end{equation}

Given the sensory data, we then aim to find the posterior over states and actions $p(\bm{x}, \bm{u}| \bm y_v, \bm y_q)$. We approximate the posterior with a variational distribution $Q(\bm{x}, \bm{u})$. We utilize the mean-field assumption ($Q(\bm{x}, \bm{u}) = Q(\bm{x})Q(\bm{u})$) and the Laplace approximation. The posterior over the state $\bm{x}$ is assumed Gaussian with mean $\bm{{\mu}}_{x}$. The posterior over actions $\bm{u}$ is also assumed Gaussian with mean $\bm{\mu}_{u}$. This results in the following expression for variational free-energy:
\begin{equation}
     F = -\ln p(\bm{\mu}_{u}, \bm{{\mu}}_{x}, \bm y_v, \bm y_q, \bm y_{\dot{q}}) + C
\end{equation}
This expression can be factorized as in equation \eqref{eq: factorization_d-aic_joint_probabilistic_distribution_ with_explicit_actions}. Assuming Gaussian model $F$ can be expanded to:
\begin{equation}
\begin{split}
    \label{eq:laplace_F_final_vector}
    F  
    &=  \frac{1}{2}(\bm{\varepsilon}_{y_q}^\top P_{y_q}\bm{\varepsilon}_{y_q}
    +  \bm{\varepsilon}_{y_{\dot{q}}}^\top P_{y_{\dot{q}}}\bm{\varepsilon}_{y_{\dot{q}}}
    + \bm{\varepsilon}_{y_v}^\top P_{y_v}\bm{\varepsilon}_{y_v}\\
    &+ \bm{\varepsilon}_{x}^\top P_{x}\bm{\varepsilon}_{x}
    + \bm{\varepsilon}_{u}^\top P_{u}\bm{\varepsilon}_{u} - \ln|P_{u}P_{y_q}P_{y_{\dot{q}}}P_{y_v}P_{x}|)
    + C,
\end{split}
\end{equation}

\noindent where $\bm{\varepsilon}_{y_q}$, $\bm{\varepsilon}_{y_{\dot{q}}}$, $\bm{\varepsilon}_{y_v}$ are the sensory prediction errors of position encoder, velocity encoder, and the visual sensor respectively. Furthermore, $\bm{\varepsilon}_{x}$ and $\bm{\varepsilon}_{u}$ are the prediction errors for the prior on the state and the control action respectively. 

\textcolor{black}{The prediction error on the state prior $\bm{\varepsilon}_{x}$ is computed considering a prediction of the state $\hat{\bm x}$ at the current time-step, which is a deterministic value. It follows that $ \bm{\varepsilon}_{x} = (\bm{\mu}_x - \hat{\bm x} )$. The prediction $\hat{\bm x} = [\hat {\bm q} \,\, \hat{ \dot{\bm q}}]^\top$ }can be computed in the same fashion as the prediction step of, for instance, a Kalman filter or of a Luenberger observer. While in such cases the prediction computation would require the knowledge of an accurate dynamic model, in this paper we assume we do not have access to that. Instead, we will  propagate forward in time the current state belief using the following simplified discrete time model:

\begin{equation}
    \label{eq:euler_integration_a}
    \hat {\bm x}_{k+1} = 
    \begin{bmatrix}
    I & I \Delta t \\
    0 & I
    \end{bmatrix}
    \bm \mu_{x,k}
\end{equation}
where $I$ represents an unitary matrix of suitable size.
This form assumes that the velocities of the joints will remain roughly constant and the position of each joint is thus computed as the discrete time integral of the velocity, using a first-order Euler scheme.
As mentioned, if one has access to a better dynamic model, that can be used instead.

Finally, the state prior now does not encode information about the target $\bm \mu_d$ (unlike in the standard AIC). The information about the target is encoded in the distribution $p(\bm u | \bm x)$. We choose this distribution to be Gaussian as well with a mean of $f^*(\bm{\mu}_x, \bm{\mu}_d)$, which is a function that steers the systems toward the target. This then results in  $\bm{\varepsilon}_{u} = (\bm{\mu}_u - f^*(\bm{\mu}_x, \bm{\mu}_d))$. The function $f^*(\bm{\mu}_x, \bm{\mu}_d)$ can be \textit{any} controller, for instance a P controller: $f^*(\bm{\mu}_x, \bm{\mu}_d)) = P (\bm{\mu}_d - \bm{\mu}_x)$. In this paper we choose the function such that it converges to a PID controller; however, other choices can be made without loss of generality. For state and actions update, first-order Euler integration is used to obtain discrete time equations.





\subsection{Definition of the observation model}
\label{subsec: observation model}
The sensory prediction errors are as in the AIC, that is $\bm\varepsilon_q = (\bm{y}_q-\bm{\mu})$, $\bm\varepsilon_{\dot{q}} = (\bm y_{\dot{q}}-\bm{\mu}')$ and $\bm\varepsilon_v = (\bm{y}_v-\bm{g_{v}}(\bm{\mu}))$ but $\bm{\mu}$ is not biased anymore towards the target $\bm{\mu}_d$. The relation between $\bm \mu$ and $\bm y$ is expressed through the generative model of the sensory input $\bm g = [\bm g_q,\ \bm g_{\dot{q}},\ \bm g_v]$. Position and velocity encoders directly measure the state. Thus $\bm g_q$ and $\bm g_{\dot{q}}$ are linear mappings between $\bm \mu$ and $\bm y$. 

To define $\bm g_v(\bm \mu)$, instead, we use a \emph{Gaussian Process Regression} (GPR) as in \cite{lanillos1}. This is particularly useful because we can model the noisy and distorted sensory input from the camera, and at the same time we can compute a closed form for the derivative of the process with respect to the beliefs $\bm \mu$, required for the state update laws. The training data is composed by a set of observations of the (planar) camera output $[\bar{\bm y}_{v_x},\ \bar{\bm y}_{v_z}]^\top$ in several robot configurations $\bar{\bm{y}}_q$. We use a squared exponential kernel $k$ of the form:
\begin{eqnarray}
\label{eq:squareexp}
\nonumber
	k(\bm{y}_{q_i},\bm{y}_{q_j}) &= \sigma_f^2 \exp\begin{pmatrix}
		-\frac{1}{2}(\bm{y}_{q_i}-\bm{y}_{q_j})^\top\Theta(\bm{y}_{q_i}-\bm{y}_{q_j})
	\end{pmatrix}\\ &+\sigma_n^2 d_{ij}
\end{eqnarray}
where $\bm{y}_{q_i},\ \bm{y}_{q_j} \in \bar{\bm y}_q$, and $d_{ij}$ is the Kronecker delta function. $\Theta$ is a diagonal matrix of hyperparameters that are fit according to the training data to obtain accurate predictions. The hyperparameters are optimized in order to maximize the marginal likelihood, see \cite{GPR_general1}. The predictions are then \cite{lanillos1}:
\begin{equation}
\begin{split}
	\bm g_v(\bm{y_{q_*}}) &= \begin{bmatrix} k(\bm{y_{q_*}},\bar{\bm{y}}_q)K^{-1}\bar{\bm y}_{v_x} \\k(\bm{y_{q_*}},\bar{\bm{y}}_q)K^{-1}\bar{\bm y}_{v_z} \end{bmatrix}\
	\\\bm g_v(\bm{y_{q_*}})' &= \begin{bmatrix} -\Theta^{-1}(\bm{y_{q_*}}-\bar{\bm{y}}_q)^\top [k(\bm{y_{q_*}},\bar{\bm{y}}_q)^\top\cdot \bm{\alpha}_x]\\ -\Theta^{-1}(\bm{y_{q_*}}-\bar{\bm{y}}_q)^\top [k(\bm{y_{q_*}},\bar{\bm{y}}_q)^\top\cdot \bm{\alpha}_z]\end{bmatrix}
\end{split}
\end{equation}

where $\cdot$ means element-wise multiplication, $K$ is the covariance matrix, $\bm{\alpha}_x = K^{-1}\bar{\bm y}_{v_x}$ and $\bm{\alpha}_z = K^{-1}\bar{\bm y}_{v_z}$.

\subsection{Estimation and control using the u-AIC}
Similar to the AIC, performing both state-estimation and control can be achieved using gradient descent on $F$. This is simpler in the case of the u-AIC since there is no need to use the chain rule when computing the control actions (see eq.~\eqref{eq:F_minimize_a}) or to consider generalized motion \eqref{eq:F_minimize_mu}: 
\begin{equation}
    \label{eq:d-AIC_F_minimize_a}
    \dot{\bm{\mu}}_u = - \kappa_{u}\frac{\partial F}{\partial \bm{\mu_u}}, \hspace{3 mm}
    \dot{\bm{{\mu}}}_x = - \kappa_{\mu}\frac{\partial F}{\partial {\bm{{\mu_x}}}},
\end{equation}

\noindent where $\kappa_{u}$ and $\kappa_{\mu}$ are the gradient descent step sizes. The gradient on control can be computed as:
\begin{equation}
    \frac{\partial F}{\partial \bm{\mu}_u}= P_u (\bm{\mu}_u - f^*(\bm{\mu}_x, \bm{\mu}_d))
\end{equation}

Setting this to zero results in the control action being equal to $f^*(\bm{\mu}_x, \bm{\mu}_d)$. If this function is chosen similar to a PID controller, then the control law of our system will converge towards that. This can be extended to richer control by modifying the probabilistic model in eq. \eqref{eq: factorization_d-aic_joint_probabilistic_distribution_ with_explicit_actions}. The only term depending on the control action now is $p(\bm u| \bm x)$; however, a prior $p(\bm u)$ can be also added. In that case the generative model would be: 
\begin{equation}
    \frac{1}{\alpha}{p(\bm{u})} {p(\bm{u}|\bm{x})} {p(\bm y_v|\bm{x}) p(\bm y_q|\bm{x}) p(\bm y_{\dot{q}}|\bm{x})} {p(\bm{x})}
\end{equation}

where $\alpha$ is a normalization constant.
This prior can then encode a feed-forward signal (open-loop control law). When computing the $\bm u$ that minimizes $F$ in that case, it would be a combination of the feed-forward signal and $f^*(\bm{\mu}_x, \bm{\mu}_d)$  depending on the value of $\Sigma_u$. This is employed in \cite{2020ICRA_baioumy}.


Since $\bm{\mu}_u$ converges to $f^*(\bm{\mu}_x, \bm{\mu}_d)$, we can achieve unbiased state estimation. In fact, this would result in $\bm\varepsilon_u = 0$ and thus the current target would not affect the the beliefs about the states. In u-AIC, the belief about the states is only affected by the predicted values and the observations from various sensor, exactly as in a Kalman filter \cite{anil_colored_noise}.


    


\subsection{Key differences between AIC and u-AIC}
The AIC considers the relationship between states and observations without explicitly modelling the control actions. The prior over state $p(\bm x)$ in equation \eqref{eq: factorization_of_coupled_AIC} thus encodes the target/goals state. In filters, such as the Kalman filter, the prior comes from a prediction step that relies on the previous state and action. In case of the AIC, the beliefs update law is essentially a filter where the prediction step is biased towards the goals state i.e. the agent normally predicts to move linearly towards the target (see eq.~\eqref{eq:genModmu}).

In the u-AIC, actions are explicitly modelled and the target state is encoded in the term $p(\bm{u}|\bm{x})$. This has two implications. First, the state is not biased towards the target since its prior $p(\bm x)$ is not. A prior encoding a prediction step will also improve the overall state-estimation accuracy. Secondly, explicit actions allow us to directly perform gradient descent on $F$ with respect to $\bm u$. This is not possible in the general AIC case and the chain rule had to be utilized (see eq.~\eqref{eq:F_minimize_a}). 

Finally, the u-AIC has guarantees regarding the stability of the controller while the AIC does not. This is out of scope of this paper and will be discussed in future work. 


\section{Fault detection, isolation, and recovery}
\label{sec:FDI}
In this section we describe how the SPE from the u-AIC can be used as residual signals, and how to perform FDI. Without loss of generality, for these derivations we will consider only the proprioceptive sensors and we discretize the continuous dynamics with first order Euler integration. According to the approximated system's dynamics in  \eqref{eq:euler_integration_a}, and expanding the second term in \eqref{eq:d-AIC_F_minimize_a} where $F$ is computed as in \eqref{eq:laplace_F_final_vector}, the state estimation law can be rewritten as:
\begin{equation}
\label{eq:estimator_mu}
\bm \mu_{k+1} = \gamma (\bm \mu_k, \bm u_k) + \Lambda(\bm g(\bm\mu_k) - \bm y_k)\\
\end{equation}

Equation \eqref{eq:estimator_mu} represents the dynamics of an estimator where stability results from the value that we obtain for diagonal matrix $\Lambda$. In \cite{anil_colored_noise} it has been shown how the free-energy principle can be used to derive stable state observers, for a linear case with coloured noise. It is important to 
note that $\gamma$ and $\Lambda$ are known expressions resulting from the partial derivatives of $F$. The term $(\bm y_k - \bm g(\bm\mu_k))$ represents the sensory prediction errors. In the u-AIC, the states of the system are steered towards the desired goal following the dynamics imposed by the controller. Therefore, the resulting system's dynamics can be represented in general as:

\begin{equation}
\label{eq:dynamics_x}
\begin{cases}
\bm x_{k+1} = \gamma (\bm x_k, \bm u_k) + \phi(\bm x_k, \bm u_k, \bm \rho_k)\\
\bm y_k = \bm x_k + \bm z_k
\end{cases}
\end{equation}
where $\bm x_k\in\mathbb{R}^n$ and  $\bm u_k\in\mathbb{R}^m$ are the state and input variables, while $\gamma (\bm x_k, \bm u_k):\mathbb{R}^n\times\mathbb{R}^m \mapsto \mathbb{R}^n$ represents the dynamics of the system in healthy conditions. $\bm y_k\in\mathbb{R}^n$ is the measurement of the full state which is affected by the presence of measurement noise $\bm z_k\in\mathbb{R}^n$.
The expressions obtained so far for the system dynamics will allow us to cast easily a model-based fault diagnosis approach in the current framework, even if the knowledge of an a-priori model is not needed. The real system can be affected by faults which are represented by the fault function $\phi(\bm x_k, \bm u_k, \bm \rho_k):\mathbb{R}^n\times\mathbb{R}^m\times \mathbb{R}^l\mapsto \mathbb{R}^n$.  The unknown parameter $\bm \rho_k$ determines the fault amplitude and shall be such that $\phi(\bm x_k, \bm u_k, \bm 0) = 0$. 

\subsection{Residual generation}
Following \cite{Rostampour2017, Rostampour2018} we rely on two assumptions:
\begin{assumption} No fault acts on the system before the fault time $k_f$, thus $\bm \rho_k = 0$ for $0\leq k< k_f$. In addition, before and and after the occurrence of a fault the variables $\bm x_k$ and $\bm u_k$ remain bounded, that is $\exists\ \mathcal{S} = \mathcal{S}^{\bm x}\times \mathcal{S}^{\bm u} \subset \mathbb{R}^n\times\mathbb{R}^m$ such that $(\bm x_k,\ \bm u_k) \in \mathcal{S}\ \forall k$.
\end{assumption} 
\begin{assumption} 
$\bm z_k$ is a random variable defined on some probability spaces ($\mathcal{Z}, \mathscr{B}, \mathbf{P}_\mathcal{Z}$), where $\mathcal{Z}\subseteq \mathbb{R}^n$. $\mathscr{B}(\cdot)$ indicates a Borel $\sigma$-algebra, and $ \mathbf{P}_\mathcal{Z}$ is the probability measures defined over $\mathcal{Z}$. Furthermore, it is not correlated and are independent from $\bm x_k$, $\bm u_k$ and $\bm \rho_k$ $\forall k$.
\end{assumption} 

We define the residuals for fault detection as the sensory prediction errors $\bm r_{k} = \bm y_k - \bm g(\bm\mu_k)$. Thus, according to \eqref{eq:dynamics_x}, \eqref{eq:estimator_mu} and following \cite{Rostampour2017, Rostampour2018}, the residuals dynamics will be given by:
\begin{equation}
\label{eq:rk+1}
    \bm r_{k+1} = \Lambda\bm r_k + \bm \delta_k + \phi(\bm x_k, \bm u_k, \bm \rho_k) 
\end{equation}

\begin{equation}
    \bm \delta_k = \gamma (\bm y_k - \bm z_k, \bm u_k) - \gamma (\bm \mu_k, \bm u_k) + \bm z_{k+1}
\end{equation}
The stochastic process $\bm \delta_k$, is the random total uncertainty which affects the residuals generation. It follows that $\bm \delta_k$ is in the probability space ($\Delta_k, \mathscr{B}(\Delta_k), \mathbf{P}_{\bm \delta_k}$) where $\Delta_k$ is obtained by letting $\bm z_k$ and $\bm z_{k+1}$ vary over $\mathcal{Z}$. Apart from simple cases, it is not possible to obtain a closed form for this set, thus numerical approximations are used instead. The residual $\bm r_{k+1}$ can be seen as a random variable in the same probability space of $\bm \delta_k$ \cite{Rostampour2017}. 

\begin{figure}[!htb]
    \centering
    \includegraphics[width=0.9\linewidth]{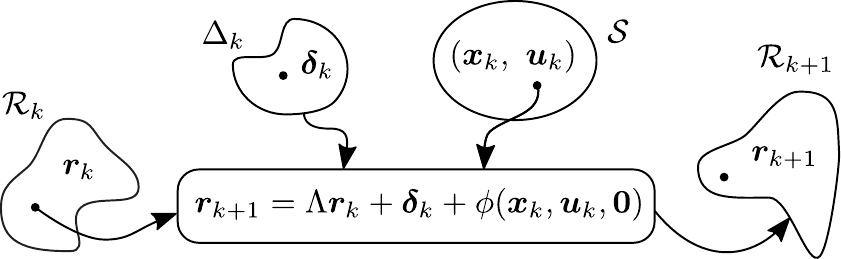}
    \caption{Healthy residual set at time $k+1$ as image obtained from the output of \eqref{eq:rk+1}.}
    \label{fig:residualset}
\end{figure}

The residual set $\mathcal{R}_{k+1}$ at the next time step $k+1$ can be seen as the image obtained by computing the output of \eqref{eq:rk+1}, where the total uncertainty $\bm \delta_k$ varies over its domain $\Delta_k$ (Fig.~\ref{fig:residualset}). Note that the healthy residuals can be characterized by setting $\bm \rho_k = 0$.

\begin{rem}
While the system dynamics $\gamma$ and the fault function $\phi$ have been introduced for analysis purposes, they do not correspond to known models of the healthy and faulty behaviours. In particular, $\gamma$ is obtained via algebraic manipulations from current free energy framework's update equations. In particular, the residual $r$ for implementing the detection and isolation logic described next can be directly computed from the current belief and measurement. Indeed, the proposed fault diagnosis approach based on the u-AIC is completely model-free.
\end{rem}

\subsection{Threshold for fault detection}
\label{sec:FD}
As in \cite{Rostampour2018}, we consider a probabilistically robust detection logic of the form:
\begin{equation}
    \label{eq:detection}
    d_M(\bm r_{k+1})\leq \frac{n}{\alpha}  \triangleq \overline{d_M}
\end{equation}
where $\overline{d_M}$ is the detection threshold, $n$ is the dimension of the residuals, $\alpha$ is a tuning parameter, and $d_M(\cdot)$ indicates the Mahalanobis distance of the residuals. In general:
\begin{equation}
    d_M(\bm r) = \sqrt{(\bm r-\bar{\bm r})^\top C_r^{-1}(\bm r-\bar{\bm r})} 
\end{equation}
where $\bar{\bm r} \triangleq \mathbb{E}[\bm r]\in \mathbb{R}^n$ and $C_r \triangleq Cov[\bm r] \in \mathbb{R}^{n\times n}$ are the expected value and covariance matrix of the residuals. According to the Multivariate Chebyshev Inequality \cite{chen2007new}:
\begin{equation} 
    Pr[ d_M(\bm r_k) \geq \frac{n}{\alpha}] \leq \alpha,\hspace{5mm} \forall \alpha \in [0, \, 1 ]
\end{equation}
This means that, in healthy conditions, the probability of a false alarm, that is $d_M$ exceeding the threshold $\bar d_M$, is upper bounded by $\alpha$.
The value $\alpha$ can be tuned to achieve the desired probabilistic robustness of the threshold. The present detection logic is equivalent to check if the residual at time step $k+1$ belongs to a time varying ellipsoid with mean $\bar{\bm r}_{k+1}$ and covariance $C_{r_{k+1}}$. In the general nonlinear case, these moments can be approximated by their sampled counterparts obtained in healthy conditions. 

\subsection{Fault isolation}
The detection policy in \eqref{eq:detection} results effective to label the system as healthy or faulty. However, it is not possible to use the same SPE generated through equation \eqref{eq:laplace_F_final_vector} for fault isolation. In fact, the free-energy is computed as a weighted sum of the squared prediction errors fusing different sensory sources. This means that when a fault occurs, its effects will propagate to all the SPE in equation \eqref{eq:laplace_F_final_vector}. We define two additional free-energies and state estimation processes which rely on different sensory sources (i.e. either proprioceptive or visual). 
\begin{equation}
\begin{split}
\label{eq:Fp}
    F_p  
    &=  \frac{1}{2}(\bm{\varepsilon_{y_q}}^\top P_{y_q}\bm{\varepsilon_{y_q}}
    +  \bm{\varepsilon_{y_{\dot{q}}}}^\top P_{y_{\dot{q}}}\bm{\varepsilon_{y_{\dot{q}}}} + \bm{\varepsilon_{x}}^\top P_{x}\bm{\varepsilon_{x}} \\
    &-\ln|P_{y_q}P_{y_{\dot{q}}}P_{x}|),
\end{split}
\end{equation}
\begin{equation}
\label{eq:Fv}
    F_v  = \frac{1}{2}(\bm{\varepsilon_{y_v}}^\top P_{y_v}\bm{\varepsilon_{y_v}}
    + \bm{\varepsilon_{x}}^\top P_{x}\bm{\varepsilon_{x}}
     - \ln|P_{y_v}P_{x}|)
\end{equation}

Doing so, we can isolate encoder and camera faults since state estimation is now done using two independent measurement sources. $F_p$ and $F_v$ are only used for fault isolation and not control. The thresholds for the SPE from \eqref{eq:Fp} and \eqref{eq:Fv} are then computed as in Sec.~\ref{sec:FD}.

\subsection{Fault recovery}
Fault recovery can be performed as in \cite{Corrado2}. 
Once a fault is detected and isolated, fault recovery is triggered. If a sensor is marked as faulty, its corresponding precision matrix is set to zero. The controller can thus exploit the sensory redundancy to a) have a better a posteriori approximation of the states, and b) to compensate for missing or wrong sensory data. Once a fault is detected and isolated, the precision matrix of the faulty sensor $P_{fs}$ is reduced to zero:
\begin{equation}
    P_{fs} = \bm 0
\end{equation}


Interestingly, the active inference framework allows also for hyper-parameters (precision) learning \cite{baioumy2020active, tutorial}. The bias in the standard AIC hindered hyper-parameters learning for fault recovery purposes, but learning meaningful precision matrices associated with sensory readings with u-AIC is now possible and will be investigated in future work. 
\section{Simulation study}
\label{sec:results}
In this section we illustrate the fault detection, isolation, and recovery strategy on a \textsc{Matlab} simulation using a 2-DOF robotic manipulator. The dynamical model of the robot is defined as \cite{SicilianoBook}:
\begin{equation}
\bm u=M(\bm q)\ddot{\bm q}+C(\bm q,\dot{\bm q})\dot{\bm q}+D\dot{\bm q}+G(\bm q) \end{equation}
where $\bm q = [q_1,\ q_2]^\top$, $\bm u(t) = [u_1,\ u_2]^\top$, $D$ is the friction coefficient matrix, $M$ is the inertia matrix, $C$ is the Coriolis matrix,  and $G$ models the effects of gravity. 
The dynamic model is only used to simulate the response to the commanded torques since the u-AIC provides a control law agnostic to the dynamical model. Thus, link's masses, friction coefficients, and other dynamical parameters are not part of the controller. The noisy sensory readings have zero-mean Gaussian noise. The standard deviation of the noise for encoders and velocity sensors is set to $\sigma_q = \sigma_{\dot{q}}  =0.001$, while the one for the camera to $\sigma_{v} = 0.01$. 



The simulation consists of controlling the robot arm for a double point-to-point motion from the starting configuration $\bm q = [-\pi/2,\ 0]\ (rad)$ to $\bm \mu_{d1} = [-0.2,\ 0.5]\ (rad)$, and then $\bm \mu_{d2} = [-0.6,\ 0.2]\ (rad)$. A fault occurs during the trajectory from $\bm \mu_{d1}$ to $\bm \mu_{d2}$, i.e. we set $k_f = 8s$. The fault can affect either the encoders or the camera. The encoder fault is such that the output related to the first joint freezes so $\bm y_q(k) = [q_1(k_f),\ q_2(k)]^\top$ for $k \geq k_f$. The fault detection and recovery of such a fault, as well as the system's response, are reported below in Fig.~\ref{fig:dmEnc} and Fig.~\ref{fig:responseEnc}:
\begin{figure}[!htb]
    \centering
    \includegraphics[width=0.90\linewidth]{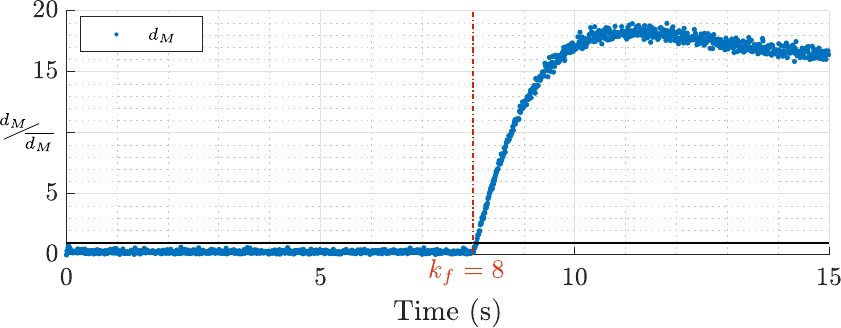}
    \caption{Normalised $d_M$ in case of encoder fault. A detection occur when the normalized value exceeds 1.}
    \label{fig:dmEnc}
\end{figure}

\begin{figure}[!htb]
    \centering
    \includegraphics[width=0.95\linewidth]{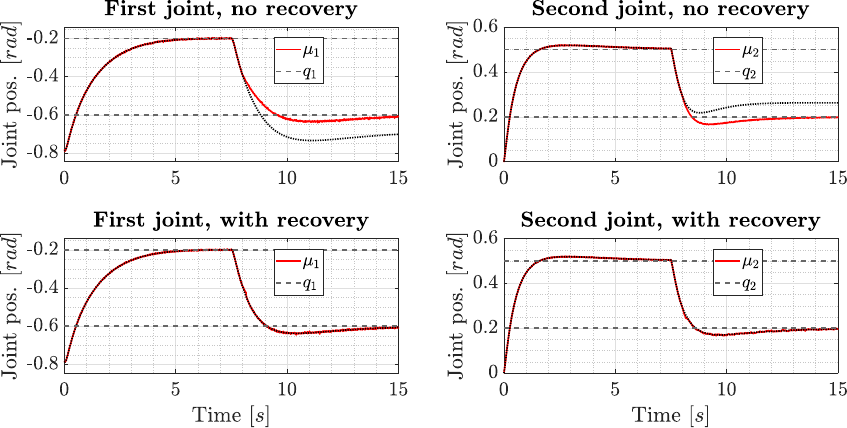}
    \caption{System's response in case of encoder fault without recovery (fixed precision matrices) and with recovery (setting $P_{y_q} = \bm 0$ after detection). The fault occurs after 8s. }
    \label{fig:responseEnc}
\end{figure}
As can be seen from Fig.~\ref{fig:responseEnc}, the system is not able to reach the set-point after the occurrence of the fault in case of no recovery. The robot arm reaches a different configuration to minimise the free-energy, which is built fusing the sensory information from the (faulty) encoders and the (healthy) camera. Recovery happens after $60\ (ms)$, and after that the faulty reading is discarded by setting zero precision. The robot arm is then able to reach the final configuration. 

\begin{figure}[!htb]
    \centering
    \includegraphics[width=0.9\linewidth]{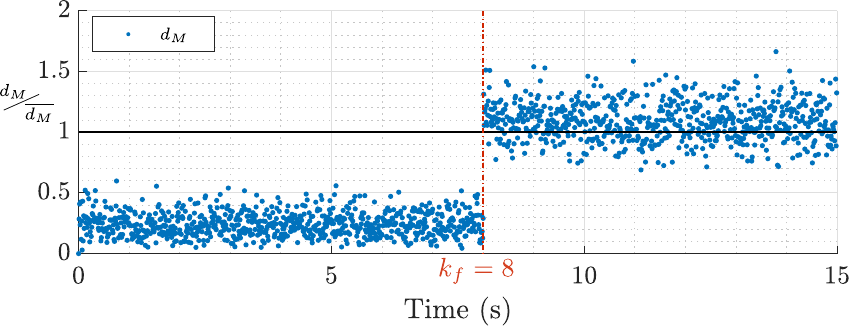}
    \caption{Normalised $d_M$ in case of camera fault.}
    \label{fig:dmCam}
\end{figure}

The other situation that we consider in this work is a camera fault. This fault is supposed to be a misalignment, due for instance to a collision. This is simulated by injecting a bias i.e. $\bm y_v(k) = \bm y_v(k) + 0.04$ for $k \geq k_f$. Note that the entity of this fault is small and comparable with the camera noise, i.e. the zero mean Gaussian noise with $\sigma_{v} = 0.01\ (m)$. The fault detection is reported in Fig.~\ref{fig:dmCam}. The system's response with and without recovery is similar to Fig.~\ref{fig:responseEnc}. 

The simulation results will be extended to real experiments, leveraging the scalability and adaptability of the active inference controller \cite{Pezzato2020}. The steady state errors ($e_{ss}$) and the $RMSE$ between beliefs and actual joint positions for the simulations are reported in Table~\ref{tab:metrics}.

\begin{table}[!htb]
\caption{Summary of the results using u-AIC}
\label{tab:metrics}
\begin{tabular}{cccccc}
\hline
\multicolumn{1}{c}{\multirow{2}{*}{\textbf{Metrics}}} & \multicolumn{2}{c}{\textbf{Encoder Fault}} & \multicolumn{2}{c}{\textbf{Camera Fault}} \\ \cline{2-5} 
\multicolumn{1}{c}{} & Recovery  & No recovery  & Recovery  & No recovery \\ \hline 
$e_{ss,\ q1}$        & $-2.7e^{-3}$  &   $0.1674$    & $-2.1e^{-4}$ & $0.0173 $        \\
$e_{ss,\ q2}$        & $-1.5e^{-3} $     & $-0.1176$    & $-2.5e^{-4}$ & $7.0e^{-3} $        \\
$RMSE_{q_1}$         &  $3.0e^{-3} $     &  $0.0896$   & $5.7e^{-4}$ & $0.0112  $        \\
$RMSE_{q_2}$         &   $2.5e^{-3}$     &   $0.0636$  & $3.5e^{-4}$ & $2.1e^{-3}  $     \\ 

\end{tabular}
\end{table}

\section{Conclusions}
\label{sec:conclusions}
In this paper we presented a new formulation of an active inference controller where the free-energy depends explicitly on the control actions. This formulation brings two general advantages. First, it allows us to remove the bias from state-estimation which affected the standard active inference controller, leading to more accurate state estimation. Second, it simplifies the definition of the control actions since it removes the need for computing partial derivatives of the free-energy. Next to that, we showed how the new formulation simplifies the use of established fault detection techniques with probabilistic robustness, which would have caused many false positives using standard active inference. The effectiveness of the approach is showcased in a simulated 2-DOF arm subject to sensory faults. Future work will investigate fault-tolerant control with stochastic decision boundaries, a formal stability proof, and experiments on a real robot manipulator.

\bibliographystyle{IEEEtran}
\bibliography{IEEEabrv,bibFile}

\end{document}